# Stochastic Rank Aggregation

**Shuzi Niu    Yanyan Lan    Jiafeng Guo    Xueqi Cheng**
Institute of Computing Technology, Chinese Academy of Sciences, Beijing, P. R. China
niushuzi@software.ict.ac.cn, {lanyanyan,guojiafeng,cxq}@ict.ac.cn

## Abstract

This paper addresses the problem of rank aggregation, which aims to find a consensus ranking among multiple ranking inputs. Traditional rank aggregation methods are deterministic, and can be categorized into explicit and implicit methods depending on whether rank information is explicitly or implicitly utilized. Surprisingly, experimental results on real data sets show that explicit rank aggregation methods would not work as well as implicit methods, although rank information is critical for the task. Our analysis indicates that the major reason might be the unreliable rank information from incomplete ranking inputs. To solve this problem, we propose to incorporate uncertainty into rank aggregation and tackle the problem in both unsupervised and supervised scenario. We call this novel framework *stochastic rank aggregation* (St.Agg for short). Specifically, we introduce a prior distribution on ranks, and transform the ranking functions or objectives in traditional explicit methods to their expectations over this distribution. Our experiments on benchmark data sets show that the proposed St.Agg outperforms the baselines in both unsupervised and supervised scenarios.

## 1 INTRODUCTION

Rank aggregation is a central problem in many applications, such as metasearch, collaborative filtering and crowdsourcing tasks, where it attempts to find a consensus ranking among multiple ranking inputs.

In literature, deterministic rank information has been utilized to solve the rank aggregation problem, and we refer these methods as *deterministic rank aggregation*. The conventional methods can be further divided into two categories: explicit and implicit methods. For explicit methods (Aslam & Montague, 2001; Cormack et al., 2009), rank information is explicitly utilized for rank aggregation, e.g., (Aslam & Montague, 2001) uses the mean rank as the score and sorts items in ascending order. While for implicit methods (Gleich & Lim, 2011; Thurstone, 1927), rank information is used implicitly, e.g., (Thurstone, 1927) defines the generative probability of pairwise preferences based on rank information and adopts the maximum likelihood procedure for aggregation.

Although rank information is crucial for the rank aggregation task, surprisingly, the explicit methods would not work well as implicit methods in practice. Our experiments on real data sets show that the implicit methods outperform explicit methods, in both unsupervised and supervised scenarios.

The root of the problem may lie in the unreliable rank information from multiple incomplete ranking inputs. Typically, the ranking inputs in rank aggregation are partial rankings. For example, in metasearch, only top search results are returned from each meta search engine with respect to its repository; in a recommender system, users only rate the items they have ever seen. The incompleteness of the partial ranking makes the ranks of items no longer reliable. Taking the recommender system as an example, we do not know whether an unseen item will be ranked higher or lower than the already rated ones for the user. As indicated by (Voorhees, 2002; Farah & Vanderpooten, 2007), the incorrect rank information will reduce the performance of any explicit method. Therefore, it is not surprising to see the failure of explicit methods based on the current unreliable rank information.

To amend this problem, we propose to incorporate uncertainty into rank aggregation and tackle the problem in both unsupervised and supervised scenario. We refer this novel rank aggregation framework as *stochastic rank aggregation* (St.Agg for short). Specifically, a prior distribution derived from pairwise contests is introduced on ranks to accommodate the unreliable rank information, since it has been proven that rank information encoded in a pairwise way will be robust (Farah & Vanderpooten, 2007). We then define the new ranking functions or objectives (in unsupervised or supervised respectively) as the expectation of those in traditional methods with respect to the rank distribution. In unsupervised scenario, the new ranking functions are used to find the final ranking list. In supervised scenario, the learning to rank technique is employed to complete the rank aggregation task. Specifically, a feature representation for each item is first designed, based on explicit features in (Aslam & Montague, 2001) and latent features in (Volkovs et al., 2012; Kolda & Bader, 2009). A gradient method is then employed to optimize the new objectives.

Our experiments on benchmark data sets show that the proposed St.Agg significantly outperforms traditional rank aggregation methods, in both unsupervised and supervised scenarios. Furthermore, we conduct experiments to demonstrate that St.Agg is more robust than traditional methods, showing the benefit of uncertainty.

In summary, we propose a novel rank aggregation framework which incorporates uncertainty of rank information. Our major contributions include: (1) the finding that partial ranking inputs in rank aggregation will make the explicit methods fail, due to unreliable rank information; (2) the definition of rank distribution based on pairwise contests, which is the basis of stochastic rank aggregation; and (3) the proposal of a unified rank aggregation framework in both unsupervised and supervised scenarios.

The rest of the paper is organized as follows. We first introduce some backgrounds on rank aggregation in Section 2, and then conduct some empirical study and analysis on why explicit methods will not work well in rank aggregation. Section 4 describes the framework of stochastic rank aggregation, including the definition of rank distribution and St.Agg in both unsupervised and supervised scenarios. Section 5 presents our experimental results and Section 6 concludes the paper.

## 2 BACKGROUNDS

In this section, we introduce some backgrounds on rank aggregation, including problem definition, previous methods and evaluation measures.

### 2.1 PROBLEM DEFINITION

In unsupervised scenario, we are given a set of $n$ items $\{x_1, \cdots, x_n\}$ and multiple ranking inputs $\tau_1, \cdots, \tau_m$ over these items. $\tau_i(x_j)$ stands for the position of $x_j$ in the ranking $\tau_i$. If the length of $\tau_i$ is $n$, $\tau_i$ is called a full ranking; otherwise, it is called a partial ranking. The goal of unsupervised rank aggregation is to find a final ranking $\pi \in \Pi$ over all the $n$ items which best reflects the ranking order in the ranking inputs, where $\Pi$ is the space of all the full ranking with length $n$. To achieve this goal, many aggregation algorithms try to optimize a similarity function $F$ between the ranking inputs $\tau_1, \cdots, \tau_m$ and the final ranking result $\pi$, while some other aggregation algorithms directly give the form of their final ranking function $f$ without explicit optimization objectives.

In supervised scenario, we are given $N$ sets of items, denoted as $D_i = \{x_1^{(i)}, \cdots, x_{n_i}^{(i)}\}, i = 1, \cdots, N$. For each item set $D_i$, a collection of ranking inputs $\tau_1^{(i)}, \cdots, \tau_{m_i}^{(i)}$ are over this set. The ground-truth labels for all items are provided in the form of multi-level ratings, such as three level ratings (2:highly relevant, 1:relevant, 0:irrelevant), denoted as $Y^{(i)} = (y_1^{(i)}, \cdots, y_{n_i}^{(i)})$. In the training process, an aggregation function $f$ of ranking inputs is learned by optimizing a sum of a similarity function $F$ on $N$ sets, and $F$ measures the similarity between these ranking inputs and the corresponding ground-truth ranking. For prediction, given any item set $D = \{x_1, \cdots, x_n\}$ and $m$ ranking inputs $\tau_1, \cdots, \tau_m$ over this set, the final ranking $\pi$ is computed by $\pi = f(\tau_1, \cdots, \tau_m)$.

### 2.2 METHODS

Previous rank aggregation methods can be divided into two categories according to the way that rank information is used: explicit and implicit rank aggregation methods. Explicit methods directly utilize rank information to define the ranking function or the objective function. While for implicit methods, other information such as pairwise preference or listwise ranking are first constructed based on the rank information, and then leveraged for rank aggregation.

#### 2.2.1 Unsupervised Aggregation Methods

Firstly, let we introduce the two kinds of methods in unsupervised scenario, respectively.

**Explicit Methods.** In unsupervised scenario, explicit methods define the ranking function as the sum of utility functions of each items's rank information, and then sort the items in descending order. The formulation is described as follows.

$$f(x_j) = \sum_{i=1}^{m} u(x_j, \tau_i), \quad (1)$$

where $u(\cdot, \cdot)$ stands for the utility function. For example, (Aslam & Montague, 2001) used the mean position as the ranking function as shown in Eq.(2), and (Cormack et al., 2009) defined the reciprocal rank as the ranking function to further emphasize the top ranked items as shown in Eq.(3).

$$f(x_j) = \sum_{i=1}^{m} u(x_j, \tau_i) = \sum_{i=1}^{m} (n - \tau_i(x_j)), \quad (2)$$

$$f(x_j) = \sum_{i=1}^{m} u(x_j, \tau_i) = \sum_{i=1}^{m} \frac{1}{C + \tau_i(x_j)}, \quad (3)$$

where $C$ is a constant, $n_i$ is the length of $\tau_i$ and $\tau_i(x_j)$ in both Eq.(2) and Eq.(3) means the position of $x_j$ in ranking $\tau_i$.

**Implicit Methods.** (Dwork et al., 2001) used a Local Kemenization procedure to approximate an optimal solution to minimize Kendall's tau distance. (Gleich & Lim, 2011) defined the difference between the pairwise preference matrix from ranking input and the aggregated preference matrix, and adopted matrix factorization for optimization. (Thurstone, 1927) and (Volkovs & Zemel, 2012) defined the similarity measure to be the generative probability of pairwise preferences and then adopted the maximum likelihood procedure for aggregation. (Guiver & Snelson, 2009) and (Lebanon & Lafferty, 2002) defined the similarity measure to be the generative probability of each ranking list and optimized this similarity function by a maximum likelihood procedure.

#### 2.2.2 Supervised Aggregation Methods

Secondly, we continue to introduce the two kinds of methods in supervised scenario, respectively.

**Explicit Methods.** For explicit methods in supervised scenario, features are first extracted from the ranking inputs, and then the rank information generated by the ranking function of these features are directly used in the objective function. After that, learning-to-rank technique are utilized for optimization. For example, (Volkovs et al., 2012) proposed to use evaluation measures as the objective function, such as NDCG (Normalized Discounted Cumulative Gain) (Järvelin & Kekäläinen, 2002), ERR (Expected Reciprocal Rank) (Chapelle et al., 2009) and RBP (Rank Biased Precision) (Moffat & Zobel, 2008). This approach is called RankAgg, and we will review their mathematical formulations in Section 2.3.

**Implicit Methods.** (Liu et al., 2007) proposed to minimize the number of disagreeing pairs between the aggregated ranking result and the ground-truth. (Volkovs & Zemel, 2012) heuristically computed the similarity between the ranking input and the ground-truth to obtain the expertise degree of the corresponding annotator. The learned weights are then used to aggregate the ranking lists for future data. (Qin et al., 2010) introduced coset-permutation distance into Plackett-Luce model for rank aggregation.

### 2.3 EVALUATION MEASURES

Rank information is explicitly used in evaluation measures for rank aggregation, such as NDCG (Järvelin & Kekäläinen, 2002), RBP (Moffat & Zobel, 2008) and ERR (Chapelle et al., 2009). We would like to express these measures as the sum of differences on each item's generated rank and ground-truth, as shown in the following equation.

$$Ev(\pi, Y) = \sum_{j=1}^{n} v(y_j, r_j), \quad (4)$$

where $Ev$ stands for any evaluation measure, $v(\cdot, \cdot)$ stands for the difference function. We give the exact forms of NDCG, RBP and ERR as follows.

$$NDCG(\pi, Y) = \sum_{j=1}^{n} \frac{g(y_j) D(r_j)}{DCG_{max}(n)}, \quad (5)$$

where $r_j$ stands for the rank of $x_j$ in the final ranking $\pi$. $g(y_j)$ is the gain function with $g(y_j) = 2^{y_j} - 1$, $D(r_j)$ is the discount function with $D(r_j) = \frac{1}{log(1+r_j)}$, and $DCG_{max}(n)$ stands for the maximum of $\sum_{j=1}^{n} g(y_j) D(r_j)$ over $\Pi$.

$$ERR(\pi, Y) = \sum_{j=1}^{n} \frac{1}{r_j} P\{\text{users stop at } r_j\}, \quad (6)$$

where the probability $P\{\text{users stop at } r_j\}$ is defined as

$$\prod_{i \in \{i | r_i < r_j\}} (1 - \frac{2^{y_i} - 1}{2^{y_{max}} - 1}) \frac{2^{y_j} - 1}{2^{y_{max}}},$$

where $y_{max}$ is the maximum of the ground-truth label.

$$RBP(\pi, Y) = \sum_{j=1}^{n} (1-p) y_j p^{r_j - 1}, \quad (7)$$

where $p \in [0, 1]$ is a constant value, for example 0.95 used in this paper.

## 3 MOTIVATION

Rank information is crucial for rank aggregation, and evaluation measures in rank aggregation are often rank dependent. However, experimental results on real data sets show that performances of explicit rank aggregation methods cannot be comparable with implicit methods in most cases. Through analysis we find that the major reason lies in the unreliable rank information used in the explicit methods directly. This motivates us to design new aggregation methods to utilize rank information in a more robust way.

## 3.1 EMPIRICAL FINDINGS

Here we conduct experiments to compare performances between the explicit methods and the implicit methods on benchmark data sets MQ2007-agg and MQ2008-agg in LETOR4.0 in both unsupervised and supervised scenario. MQ2007-agg contains 1692 queries with 21 ranking inputs per query on average and MQ2008-agg contains 784 queries with 25 ranking inputs per query on average. In both data sets, three level relevance judgment per document is provided as the ground-truth.

In unsupervised scenario, we choose BordaCount (Aslam & Montague, 2001), RRF (Cormack et al., 2009) as the baselines of explicit method, and SVP (Gleich & Lim, 2011), MPM (Volkovs & Zemel, 2012) and PlackettLuce (Guiver & Snelson, 2009) as the baselines of implicit methods. In supervised scenario, we choose RankAgg (RankAgg$_{NDCG}$, RankAgg$_{ERR}$, RankAgg$_{RBP}$) (Volkovs et al., 2012) as the baselines of explicit methods, and CPS (Qin et al., 2010) and $\theta$-MPM (Volkovs & Zemel, 2012) as the baselines of implicit methods, where the feature mapping method used in RankAgg in supervised scenario is Borda Count (Aslam & Montague, 2001).

The experimental results are shown in Figure 1, where Bestinput in Figure 1(a) stands for the method that directly utilizes the best ranking input in terms of evaluation measures as the output. It is obvious that implicit aggregation methods outperform explicit methods in most cases in both unsupervised scenario (Figure 1(a)) and supervised scenarios (Figure 1(b)), especially on MQ2007-agg.

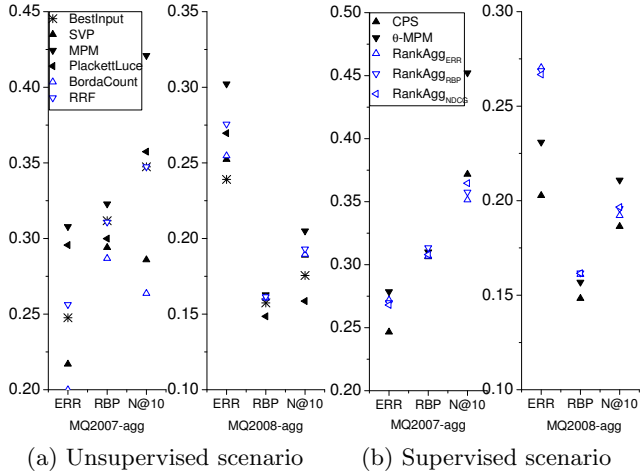

(a) Unsupervised scenario  (b) Supervised scenario

Figure 1: Performance Comparison between Existing Explicit and Implicit Methods

## 3.2 ANALYSIS

The contradiction between the experimental results and the intuition inspires us to revisit these explicit methods. The ranking inputs for aggregation, i.e. inputs in the data sets of MQ2007-agg and MQ2008-agg, are typically partial rankings. The incompleteness of the partial ranking makes the ranks of items no longer reliable, which causes the failure of these explicit methods according to (Voorhees, 2002; Farah & Vanderpooten, 2007).

Taking Borda Count and an example as follows to show the reason. $\tau_{1D} = a \succ b, \tau_{2D} = b \succ c, \tau_{3D} = c \succ d$ are three incomplete rankings over $D = \{a, b, c, d\}$, and the ground-truth ranking is $a \succ b \succ c \succ d$. Borda Count ranks the items by the mean positions of all the three lists. With assumption that $\tau_{1D}(a)=1, \tau_{1D}(b)=2, \tau_{2D}(b)=1, \tau_{2D}(c)=2, \tau_{3D}(c)=1, \tau_{3D}(d)=2$, the final ranking obtained from Borda Count in unsupervised scenario is $b \succ c \succ a \succ d$ or $c \succ b \succ a \succ d$, which is far from the ground-truth ranking. In supervised scenario, these unreliable rank positions used to calculate the Borda Count score are employed as features directly. These unreliable features lead to the failure of learning to rank algorithms.

Through further analysis, we find that the above failure is caused by unreliable rank information generated from partial ranking inputs. Specifically, the ranks of missing items are taken as NULL, which would not appear in the computation, and make the positions of remaining items no longer reliable. For example, $c$ and $d$ are missing in $\tau_{1D}$, and thus the ranks of $a$ and $b$ are taken as 1 and 2, respectively. However, these ranks do not reflect the true absolute ranks of $a$ and $b$ if $c$ and $d$ are taken into consideration. In a word, the incomplete ranking inputs result in unreliable rank information, which leads to the failure of explicit aggregation methods. This motivates us to take the uncertainty of rank information into consideration and design new rank aggregation methods.

## 4 STOCHASTIC RANK AGGREGATION

Through analysis in section 3.2, it is important for aggregation methods to be able to accommodate the uncertainty of the rank information. To achieve this goal, we treat rank as a random variable and design a novel rank aggregation framework which transforms the ranking functions or objectives into the expectations over its distribution in both unsupervised and supervised scenario, namely St.Agg.

## 4.1 INCORPORATING UNCERTAINTY INTO RANKS

In section 3.1 we observe that these implicit methods with pairwise preferences as inputs outperform explicit methods in most cases. Inspired by this observation, we define the random variable of rank with respect to a ranking input $\tau$ as the result of pairwise contests, described as follows:

$$R(x_j, \tau) = \sum_{i=1, i \neq j}^{n} I(x_i \succ_\tau x_j), \quad (8)$$

where $R(x_j, \tau)$ is the rank of item $x_j$ in $\tau$, and $I(x_i \succ_\tau x_j)$ stands for the event that item $x_j$ is beaten by item $x_i$. We assume that the pairwise contest between $x_i$ and $x_j$ follows a Bernoulli trail with probability $p(x_i \succ_\tau x_j)$ that $x_j$ is beaten by $x_i$. Therefore, the rank $R(x_j, \tau)$ is a Binomial-like random variable, equal to the number of successes of $n-1$ Bernoulli trials.

Furthermore, the rank distribution for each item $x_j$ can be computed from $\tau$ in the following process.

(1) If we only have item $x_j$, its rank will be 0 (the best rank). Thus the initialization distribution of item $x_j$ is defined as:

$$P^{(1)}(R(x_j, \tau) = r) = \delta(r) = \begin{cases} 1, & r = 0 \\ 0, & r > 0 \end{cases} \quad (9)$$

(2) Each time we add a new item $x_i$, the rank will get one larger if $x_j$ is beaten by $x_i$ and it will stay unchanged otherwise. Therefore the rank distribution for item $x_j$ will be updated by the following recursive relation:

$$\begin{aligned} P^{(t)}&(R(x_j, \tau) = r) \\ &= P^{(t-1)}(R(x_j, \tau) = r-1) p(x_i \succ_\tau x_j) \\ &+ P^{(t-1)}(R(x_j, \tau) = r)(1 - p(x_i \succ_\tau x_j)). \end{aligned} \quad (10)$$

(3) After $n-1$ iterations, $P^{(n)}(R(x_j, \tau) = r)$ will be the final distributions $P(R(x_j, \tau) = r)$.

Based on the definition of rank distribution above, the ranking function or objective function used in an explicit method can be turned into a new function by taking the expectation over this rank distribution, which will be used in the new rank aggregation method, described in the following section.

## 4.2 UNSUPERVISED ST.AGG

First we introduce the specific form of pairwise probability $p(x_i \succ_\tau x_j)$ for the computation of rank distribution $P(R(x_j, \tau) = r)$ in unsupervised scenario.

The expectations of ranking functions in explicit methods are then computed over the rank distribution, and used to obtain the final aggregated ranking. We denote our unsupervised St.Agg as St.Agg$_f$ for different ranking functions.

### 4.2.1 Definition of Pairwise Probability

Inspired by the robust rank difference information (Farah & Vanderpooten, 2007) underlying in the ranking list, we define $p(x_i, x_j)$ as the function of rank difference between $x_i$ and $x_j$ in the full ranking $\tau_{D_\tau}$ over the subset $D_\tau \subseteq D$ such as $\frac{|\tau_{D_\tau}(x_i) - \tau_{D_\tau}(x_j)|}{n}$. We would like to give an example of the definition of the pairwise probability $p(x_i \succ_\tau x_j)$ satisfying $p(x_i \succ_\tau x_j) + p(x_j \succ_\tau x_i) = 1$ as below:

$$\begin{cases} \min\{p(x_i,x_j), 1-p(x_i,x_j)\}, & \text{if } \tau_{D_\tau}(x_i) > \tau_{D_\tau}(x_j) \\ \max\{p(x_i,x_j), 1-p(x_i,x_j)\}, & \text{if } \tau_{D_\tau}(x_i) < \tau_{D_\tau}(x_j) \\ 0.5, & \text{otherwise} \end{cases} \quad (11)$$

Incorporating Eq.(11) to the recursive process Eq.(10), we can obtain the probability distribution over ranks $P(R(x_j, \tau) = r)$.

### 4.2.2 Expectations

As mentioned in section 2, the ranking function $f(x_j)$ in explicit methods is a sum of the utility function of rank information of a certain item from multiple ranking inputs. Thus we can simply calculate the expectation of the ranking function $f_s(x_j)$ over the rank distribution proposed above.

For example, the new ranking function $f_s(x_j)$ in St.Agg$_{BC}$ and St.Agg$_{RRF}$ can be obtained by incorporating rank distribution $P(R(x_j, \tau_i) = r)$ into the mean position function in Eq.(2) and the reciprocal rank function in Eq.(3) respectively, as shown below.

$$f_s(x_j) = \frac{1}{m} \sum_{i=1}^{m} \sum_{r=0}^{n-1} (n-r) P(R(x_j, \tau_i) = r), \quad (12)$$

$$f_s(x_j) = \sum_{i=1}^{m} \sum_{r=0}^{n-1} \frac{P(R(x_j, \tau_i) = r)}{r + C}. \quad (13)$$

where $BC$ is short for Borda Count. For such new aggregation methods, the final ranking is obtained by sorting in descending order of $f_s(x_j)$.

## 4.3 SUPERVISED ST.AGG

In supervised scenario, we utilize the state-of-the-art learning framework for the optimization problem in rank aggregation like RankAgg$_F$ (Volkovs et al., 2012) mentioned in the section 2.2. Similarly we denote our supervised St.Agg as St.Agg$_F$ for different definitions of optimization functions.

Firstly, the specific form of pairwise probability $p(x_i \succ_\pi x_j)$ for the computation of rank distribution $P(R(x_j, \pi) = r)$ is defined based on the aggregated ranking $\pi$. Then the optimization objectives in our supervised St.Agg can be defined as the expectation of these objectives over rank distributions $P(R(x_j, \pi) = r)$. To solve this aggregation problem by a learning procedure, a proper feature mapping is first designed for representation, and then a gradient-based optimization method is adopted to learn the ranking function $f$.

### 4.3.1 Definition of the Pairwise Probability

Pairwise probability is defined on the basis of the ranking function $f$. Therefore, we define the score of each item $x_i$ as a normal random variable denoted as $s_i$ with expectation $f(x_i)$ and variance $\sigma^2$, i.e. $s_i \sim \mathcal{N}(f(x_i), \sigma^2)$.

Similar to the definition in unsupervised scenario, the pairwise probability can be defined as the function of score difference to reflect that the larger the score difference between $x_i$ and $x_j$, the more probable that $x_j$ is defeated by $x_i$. Thus $p(x_i \succ_\pi x_j)$ can be defined as $P(s_i - s_j > 0)$, where $s_i - s_j \sim \mathcal{N}(f(x_i) - f(x_j), 2\sigma^2)$. The computation of $p(x_i \succ_\pi x_j)$ can be taken as the following Eq.(14).

$$p(x_i \succ_\pi x_j) = \int_0^{+\infty} \frac{1}{2\sigma\sqrt{\pi}} e^{-\frac{(x-(f(x_i)-f(x_j)))^2}{4\sigma^2}} dx \qquad (14)$$

Applying the specific form of $p(x_i \succ_\pi x_j)$ in Eq.(14) into the recursive computation of $P(R(x_j, \pi) = r)$ in Eq.(10), the Binomial-like distribution $P(R(x_j, \pi) = r)$ are computed. For the convenience of computing the derivation, the Binomial-like distribution can be approximated by the normal distribution with means $\sum_{i=1, i \neq j}^{n} p(x_i \succ_\pi x_j)$ and variance $\sum_{i=1, i \neq j}^{n} p(x_i \succ_\pi x_j)(1 - p(x_i \succ_\pi x_j))$.

### 4.3.2 Expectations

Objective functions in explicit methods mentioned in section 2.2 can be expressed as the sum of difference functions, such as ERR in Eq.(6), RBP in Eq.(7) and NDCG in Eq.(5). Incorporating $P(R(x_j, \pi) = r)$ into these objectives, the expectation of them can be easily obtained by taking the expectation of the difference functions $v(\cdot, \cdot)$ over the rank distribution, denoted as ERR$_s$, RBP$_s$ and NDCG$_s$. The general form $Ev_s(\pi, Y)$ and these three measures are listed below.

$$Ev_s(\pi, Y) = \sum_{j=1}^{n} \sum_{r=0}^{n-1} v(y_i, r) P(R(x_j, \pi) = r),$$

$$\text{ERR}_s(\pi, Y) = \sum_{j=1}^{n} \sum_{r=0}^{n-1} \frac{P(\text{users stop at } r) P(R(x_j, \pi) = r)}{r+1},$$

$$\text{RBP}_s(\pi, Y) = (1-p) \sum_{j=1}^{n} \sum_{r=0}^{n-1} y_j p^r P(R(x_j, \pi) = r),$$

$$\text{NDCG}_s(\pi, Y) = \frac{\sum_{j=1}^{n} \sum_{r=0}^{n-1} g(y_j) D(1+r) P(R(x_j, \pi) = r))}{DCG_{max}(n)}.$$

### 4.3.3 Feature-based Learning Framework

The remaining question is how to optimize these expected objectives in a feature-based learning framework. Stage I is to design a better feature mapping for item representation; Stage II is the learning process of the supervised St.Agg, i.e. St.Agg$_F$.

*Stage I: Feature mapping.* In literature, feature mapping techniques for rank aggregation can be classified into three groups in terms of information used in the feature extraction process.

(1) *Features in terms of user-item relation.* Borda Count is such a natural feature, which aims to assign an item a relevance score per ranking input according to its position in this ranking input only. For example $nBC(x_i, \tau_i) = \frac{n - \tau_i(x_i)}{n}$, so $\Psi_{BF}(i) = [nBC(x_i, \tau_1), \cdots, nBC(x_i, \tau_m)]$.

(2) *Features in terms of both user-item and item-item relations.* Maksims et al. (Volkovs et al., 2012) proposed a transformation from all the ranking inputs into latent feature representations for each item based on SVD factorization. Each ranking input $\tau_i$ can be transformed into a pairwise preference based matrix denoted as $P_i$. Each matrix $P_i$ can be approximated by rank-$p$ singular vector decomposition $P_i \approx U_i \Sigma_i V_i'$. Therefore, each item $x_i$ can be represented as a SVD-based feature vector from $m$ ranking inputs, $\Psi_{MF}(i) = [U_1(i,:), diag(\Sigma_1), V_1(i,:), \cdots, U_m(i,:), diag(\Sigma_m), V_m(i,:)]$.

(3) *Features in terms of all of the three relations.* Tensor factorization method can take the item-item, item-user, user-user relations into consideration (Hong et al., 2012). In this paper, we use CanDecomp/Parafac (CP) decomposition (Kolda & Bader, 2009) for tensor factorization due to the nice property that it has a unique solution of decomposition, which provides a theoretical guarantee to get a stable solution. Specifically, the item-item-user tensor $T$ with $T(:,:,i) = P_i$ is factorized as $T = \sum_{j=1}^{p} \lambda_j U_{:,j} V_{:,j} W_{:,j}$. Therefore CP-based feature vector for item $x_i$ is represented as $\Psi_{TF}(i) = [U(i,:), V(i,:)]$.

*Stage II: Gradient-based learning algorithm.* Suppose $f$ is a linear model with parameter $w$. Here we use gradient method for the optimization of these expected objectives such as $\text{ERR}_s$, $\text{RBP}_s$ and $\text{NDCG}_s$. The gradients of these expected objectives are computed as below.

$$\sum_{j=1}^{n}\sum_{r=0}^{n-1}\frac{\partial Ev_s(\pi,Y)}{\partial P(R(x_j,\pi)=r)}\frac{\partial P(R(x_j,\pi)=r)}{\partial w} \quad (15)$$

For $\text{ERR}_s$, $\text{RBP}_s$ and $\text{NDCG}_s$, the only difference of their partial derivatives lies in the first part of derivatives in the Eq.(15), which can be easily derived as follows.

$$\frac{\partial \text{ERR}_s(\pi,Y)}{\partial P(R(x_j,\pi)=r)} = \frac{P(\text{users stop at } r)}{r+1},$$

$$\frac{\partial \text{RBP}_s(\pi,Y)}{\partial P(R(x_j,\pi)=r)} = (1-p)y_j p^r,$$

$$\frac{\partial \text{NDCG}_s(\pi,Y)}{\partial P(R(x_j,\pi)=r)} = \frac{g(y_j)D(r)}{DCG_{max}(n)}.$$

## 5 EXPERIMENTS

In this section we compare the performance of our aggregation methods St.Agg with traditional methods in terms of ERR, RBP and NDCG on two benchmark aggregation data sets, i.e. MQ2007-agg and MQ2008-agg in LETOR4.0. In unsupervised scenario, the ground-truth is only used for evaluation; In supervised scenario, the ground-truth is employed for both training and evaluation. Finally we make a detailed analysis on the robustness of St.Agg.

### 5.1 EFFECTIVENESS OF UNSUPERVISED ST.AGG

As summarised in Section 2, the baseline methods fall into two groups in unsupervised scenario, i.e. the implicit group including Markov Chain based methods denoted as MCLK (Dwork et al., 2001), SVP (Gleich & Lim, 2011), MPM (Volkovs & Zemel, 2012) and Plackett-Luce (Guiver & Snelson, 2009), and the explicit group including Borda Count (Aslam & Montague, 2001) and RRF (Cormack et al., 2009). We implement two unsupervised St.Agg methods including St.Agg$_{BC}$ and St.Agg$_{RRF}$.

We use the standard partition in LETOR4.0. For parameter setting, we choose the parameters when a method reaches its best performance on validation set. For example, parameter $C$ of RRF is set to 40 on MQ2007-agg and 10 on MQ2008-agg. The learning rate is 0.1 and precision is 0.01 for SVP on both data sets.

The experimental results are shown in Table 1. Firstly, we make a comparison between the explicit methods and St.Agg in terms of NDCG@5, NDCG@10, ERR and RBP. We can see that St.Agg$_{BC}$ and St.Agg$_{RRF}$ have obvious advantage over the explicit methods Borda Count and RRF in terms of all the measures. For example, the highest performance improvement of St.Agg$_{BC}$ is 79.7% in terms of NDCG@5 on MQ2007-agg compared with Borda Count. The highest performance improvement of St.Agg$_{RRF}$ is 32.3% in terms of NDCG@5 compared with RRF. Similar results can be found on MQ2008-agg. It demonstrates that explicit rank aggregation methods can be significantly improved by incorporating uncertainty into rank aggregation.

Secondly, we make a comparison between the implicit methods and St.Agg in terms of NDCG@5, NDCG@10, ERR and RBP. We can see that St.Agg is consistently better even than the best implicit method (MPM). Compared with MPM, St.Agg$_{RRF}$ achieves 5.2% higher in terms of NDCG@5 on MQ2007-agg. In terms of NDCG@10, our St.Agg$_{BC}$ performs 4.8% better than the best implicit method (MPM) on MQ2007-agg. Similar results can be found on the MQ2008-agg. Therefore, we conclude that explicit rank aggregation methods can outperform the implicit methods after incorporating uncertainty into rank aggregation.

In summary, St.Agg$_f$ with expected ranking function can improve the performance compared with the explicit methods which utilize rank information directly for aggregation. It also turned out to be better than the state-of-art implicit aggregation methods on both data sets in terms of all the evaluation measures. Therefore, our proposal to incorporate uncertainty into rank aggregation, i.e. stochastic rank aggregation, is significant for this task.

### 5.2 EFFECTIVENESS OF SUPERVISED ST.AGG

Similarly in supervised scenario, our baseline methods fall into two categories: (1) implicit rank aggregation methods including CPS (Qin et al., 2010) and $\theta$-MPM (Volkovs & Zemel, 2012); and (2) explicit rank aggregation methods including methods mentioned in (Volkovs et al., 2012), denoted as RankAgg.

Both RankAgg and St.Agg are in a feature-based learning framework. Therefore, it is also a key problem to design a feature mapping for each item. In this paper, three mappings are adopted: (1) Borda Feature $\Psi_{BF}$; (2) SVD-based Features $\Psi_{MF}$; and (3) CP-based Features $\Psi_{TF}$. RankAgg and St.Agg under these different mappings are denoted as RankAgg($\Psi_{BF}$), RankAgg($\Psi_{MF}$), RankAgg($\Psi_{TF}$) and St.Agg($\Psi_{BF}$), St.Agg($\Psi_{MF}$) St.Agg($\Psi_{TF}$), respectively.

We use the standard partition in LETOR4.0, and em-ploy five-fold cross validation to evaluate the performance of each methods. For gradient descent procedure in CPS (Qin et al., 2010), RankAgg (Volkovs et al., 2012) and our St.Agg, learning rate is chosen from $10^{-1}$ to $10^{-6}$ when the best performance is achieved on the validation set with the maximal number of iterations is 500. An additional parameter $\sigma$ needs to be tuned from $10^{-1}$ to $10^{-4}$ for St.Agg.

Table 1: Performance Comparison under Unsupervised Methods on MQ2007-agg and MQ2008-agg. All the results with bold type are significantly better than the baseline methods ($p$-value $<$ 0.05).

| (a) MQ2007-agg | | | | | (b) MQ2008-agg | | | | |
|---|---|---|---|---|---|---|---|---|---|
| Methods | N@5 | N@10 | ERR | RBP | Methods | N@5 | N@10 | ERR | RBP |
| BestInput | 0.3158 | 0.3474 | 0.2476 | 0.3117 | BestInput | 0.3813 | 0.1756 | 0.2391 | 0.1574 |
| MCLK | 0.2098 | 0.2450 | 0.1905 | 0.2798 | MCLK | 0.3402 | 0.1431 | 0.2055 | 0.1449 |
| SVP | 0.2582 | 0.2859 | 0.2169 | 0.2941 | SVP | 0.4004 | 0.1890 | 0.2523 | 0.1606 |
| Plackett-Luce | 0.3462 | 0.3574 | 0.2957 | 0.2999 | Plackett-Luce | 0.3737 | 0.1586 | 0.2696 | 0.1485 |
| MPM | 0.3986 | 0.4210 | 0.3078 | 0.3229 | MPM | 0.4283 | 0.2050 | 0.3023 | 0.1628 |
| BordaCount | 0.2325 | 0.2637 | 0.2000 | 0.2868 | BordaCount | 0.4052 | 0.1895 | 0.2547 | 0.1607 |
| RRF | 0.3172 | 0.3474 | 0.2563 | 0.3108 | RRF | 0.4239 | 0.1931 | 0.2756 | 0.1615 |
| St.Agg$_{BC}$ | 0.4179 | 0.4384 | 0.3197 | **0.3347** | St.Agg$_{BC}$ | **0.4515** | 0.2151 | 0.3021 | 0.1671 |
| St.Agg$_{RRF}$ | **0.4195** | **0.4404** | **0.3199** | 0.3346 | St.Agg$_{RRF}$ | 0.4512 | **0.2157** | **0.3028** | **0.1673** |

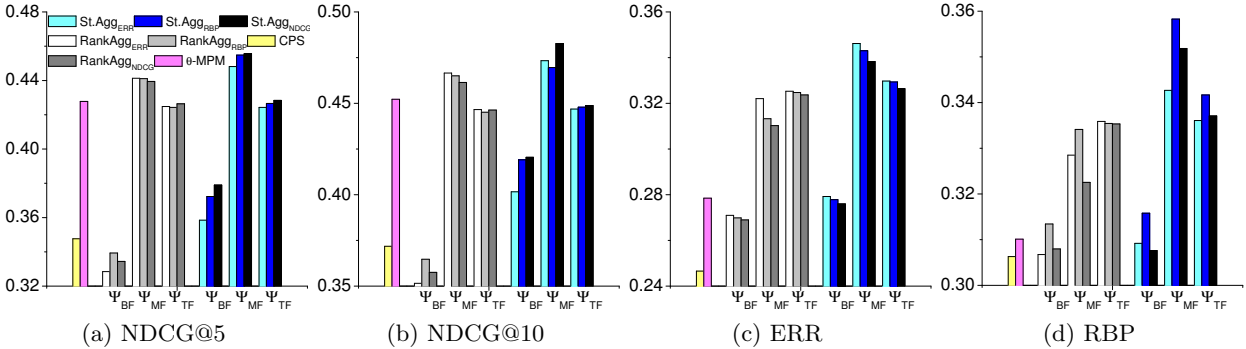

Figure 2: Performance Comparison of Supervised Aggregation Methods on MQ2007-agg

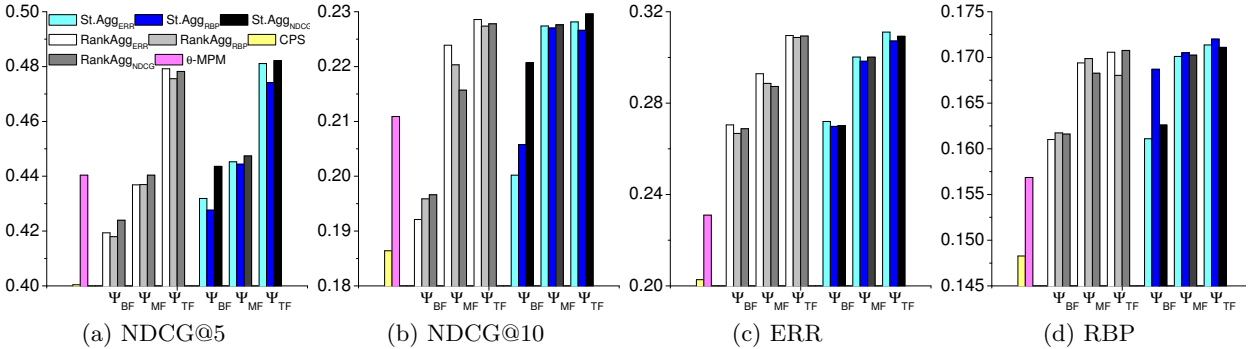

Figure 3: Performance Comparison of Supervised Aggregation Methods on MQ2008-agg

### 5.2.1 Comparison of Aggregation Methods

We first compare the performances of different methods, and the experimental results are listed in Figure 2 and Figure 3. We can see that St.Agg outperforms RankAgg consistently in terms of NDCG@10, ERR and RBP under each feature mapping ($p$-value$<$ 0.05). For example in Figure 2 on MQ2007-agg under $\Psi_{MF}$, the improvement of St.Agg$_{NDCG}$ over RankAgg$_{NDCG}$ is 4.6% in terms of NDCG@10, the improvement of St.Agg$_{ERR}$ over RankAgg$_{ERR}$ is 7.5% in terms of ERR, and the improvement of St.Agg$_{RBP}$ over RankAgg$_{RBP}$ is 7.2% in terms of RBP.

Compared with the best of implicit aggregation methods ($\theta$-MPM) in Figure 2 on MQ2007-agg, St.Agg performs consistently better under any feature mapping ($p$-value$< 0.01$). Specifically, the improvement of St.Agg$_{NDCG}(\Psi_{MF})$ over $\theta$-MPM is 6.73% in terms of NDCG@10, the improvement of St.Agg$_{ERR}(\Psi_{MF})$ over $\theta$-MPM is 24.3% in terms of ERR and the improvement of St.Agg$_{RBP}(\Psi_{MF})$ over $\theta$-MPM is 15.5% in terms of RBP. Similar results can be found on MQ2008-agg as shown in Figure 3.

To sum up, we can see that our proposed St.Agg can significantly outperform all these baselines in terms of NDCG, ERR and RBP.

### 5.2.2 Feature Mapping Comparison

We further compare the effectiveness of different feature mappings. From the results in Figure 2, we can see that $\Psi_{MF}$ is the best among all the three mappings on MQ2007-agg ($p$-value$< 0.01$). For example, performances on $\Psi_{MF}$ are consistently better than the other two mappings $\Psi_{BF}$ and $\Psi_{TF}$ for both St.Agg$_{NDCG}$ and RankAgg$_{ERR}$. In Figure 3, obviously $\Psi_{TF}$ is the best among all the three mappings on MQ2008-agg ($p$-value$< 0.01$). For example, performances on $\Psi_{TF}$ are consistently better than that on $\Psi_{MF}$ and $\Psi_{BF}$ for both St.Agg$_{NDCG}$ and RankAgg$_{NDCG}$.

Through above analysis, $\Psi_{MF}$ is best on MQ2007-agg and $\Psi_{TF}$ is the best on MQ2008-agg. Since MQ2008-agg is much smaller and noisier than MQ2007-agg, our experimental results agreed with the previous findings that feature mappings based on tensor factorization will be more robust for sparsity and noise (Kolda & Bader, 2009).

### 5.3 ROBUSTNESS ANALYSIS OF ST.AGG

It is important to consider the robustness of rank aggregation methods. Here robustness means that the comparative performance will change little along with different ranking inputs, as defined in (Carterette & Petkova, 2006). Considering the computational efficiency, here we only take unsupervised St.Agg for example. With the number of ranking inputs from 5 to 20 with a step 5, we randomly choose the ranking inputs from the whole data sets 20 times. Each point in Figure 4 depicts the average NDCG@5 obtained on these 20 results.

It is obvious that NDCG@5 of St.Agg keeps high above all these explicit and implicit methods as the number

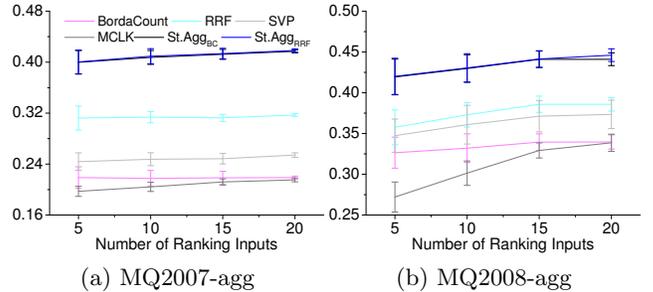

(a) MQ2007-agg  (b) MQ2008-agg

Figure 4: Robustness Analysis of St.Agg

of ranking inputs increases in the data set. Therefore, St.Agg is very robust to the change of ranking inputs by incorporating uncertainty into ranks. In addition, the performance of each method tends to be stable with the increase of the number of ranking inputs since its variance is smaller and smaller.

## 6 CONCLUSION

In this paper, we propose a novel rank aggregation framework to incorporate uncertainty to this task, namely stochastic rank aggregation (i.e. St.Agg).

We give some empirical results and analysis to show that unreliable rank information from incomplete ranking inputs will make the approaches directly using rank information fail in practice. To tackle this problem, we propose to treat rank as a random variable and define the distribution by pairwise contests. After that, a novel rank aggregation framework in both unsupervised and supervised scenario is proposed, which takes the expectations of traditional ranking functions or objective functions for optimization. Finally, our extensive experiments on benchmark data sets show that the proposed St.Agg is better in terms of both effectiveness and robustness.

For future work, it is interesting to investigate how to incorporate uncertainty to implicit methods, and whether there are better ways to define the rank distribution.

**Acknowledgements**

This research work was funded by the National Natural Science Foundation of China under Grant No. 61232010, No. 61203298, No. 61003166 and No. 60933005, 863 Program of China under Grants No. 2012AA011003, and National Key Technology R&D Program under Grant No. 2011BAH11B02, No. 2012BAH39B02, No. 2012BAH39B04. We also wish to express our thanks to Yahoo!Research for providing the Yahoo!Rand dataset to us.